# Efficient Gradient Estimation for Motor Control Learning

**Gregory Lawrence**
Computer Science Division
U.C. Berkeley
gregl@cs.berkeley.edu

**Noah Cowan**
Dept. of Mechanical Engineering
Johns Hopkins University
ncowan@jhu.edu

**Stuart Russell**
Computer Science Division
U.C. Berkeley
russell@cs.berkeley.edu

## Abstract

The task of estimating the gradient of a function in the presence of noise is central to several forms of reinforcement learning, including policy search methods. We present two techniques for reducing gradient estimation errors in the presence of observable *input* noise applied to the control signal. The first method extends the idea of a reinforcement baseline by fitting a local model to the response function whose gradient is being estimated; we show how to find the response surface model that minimizes the variance of the gradient estimate, and how to estimate the model from data. The second method improves this further by discounting components of the gradient vector that have high variance. These methods are applied to the problem of motor control learning, where actuator noise has a significant influence on behavior. In particular, we apply the techniques to learn locally optimal controllers for a dart-throwing task using a simulated three-link arm; we demonstrate that the proposed methods significantly improve the response function gradient estimate and, consequently, the learning curve, over existing methods.

## 1 INTRODUCTION

From its earliest days, reinforcement learning has been concerned with, among others, motor control problems. Analytical solutions to motor control problems are often elusive because of nonlinear dynamics and high-dimensional state spaces. Another challenge is the noise inherent to real systems. For biological systems in particular, variation in the *actual* force exerted by a muscle compared to the "commanded" force is critical to performance [3]. Although noise has often been considered little more than a nuisance for mathematical treatments of control systems, it is now believed to be a major determinant of the actual motor control strategies employed by animals and humans. The reason is quite simple: Motor control systems are highly redundant (i.e., have more degrees of freedom than required for most tasks) and in addition admit virtually infinite variation of forces over time; thus, a noise-free system may admit a high-dimensional continuous manifold of perfect solutions to a problem such as throwing a dart at a bullseye. On the other hand, once noise is introduced, some of these "perfect" strategies may prove to be extremely "fragile," whereas others may be "robust."

Working from biologically reasonable assumptions of approximately linear dynamics and multiplicative noise (i.e. noise proportional to torque exerted), Wolpert [3] found a unique optimal solution for eye saccades that closely matches observed motion profiles. Todorov and Jordan [8] derived optimal linear feedback controllers and observers for linear systems under multiplicative noise and were able to explain a number of qualitative features of biological motor control as strategies for minimizing the impact of noise on achievement of the objective.

For many problems, linearization may not work. In this paper, we consider a reinforcement learning approach based on *policy search*, i.e. directly modifying the parameters of a control policy based on observed rewards. The key challenge involves estimating the *gradient* of the expected total reward with respect to the policy parameters, given noisy training data. Perhaps the most straight-forward method is to calculate the empirical gradient at a given nominal policy based on evaluation of nearby nominal policies; this tends to require a great deal of data because it requires comparing noisy estimates of very similar quantities. Williams' REINFORCE algorithm [10] shows how to estimate the gradient at a point in parameter space using only training samples generated by the corresponding nominal policy. We describe and illustrate this method in Section 2.

Williams points out that estimator variance can be reduced by subtracting a *baseline* from the total observed reward in each training sample, and Weaver and Tao [9] derive an expression for the constant baseline that minimizes the variance of the gradient estimate. In Section 3, we view the constant baseline as a trivial *response surface model* [5] for the value of the initial state, as a function of the policy parameters, given the current nominal policy. By extending this idea to linear models, we obtain a substantial reduction in variance. Furthermore, we show that for sequential problems, it is possible to get still more variance reduction by reweighting the gradient contributions from each time step in the trial to reduce the impact of nearly-deterministic steps. Sections 4 and 5 demonstrate these results for dart-throwing with a three-link arm.

Our algorithms assume observable input noise from a known distribution. These assumptions can be relaxed somewhat (Section 6). They can also be strengthened in the simulator setting, where the "random" noise perturbations can be fixed in advance. PEGASUS [6] takes advantage of this by reusing the same random number sequence for each set of trials conducted at each nominal policy. In this way, the problem of statistical comparisons obscuring the difference in value of two nearby policies is eliminated. We compare our algorithms with PEGASUS in Section 5, and comment on possible synergies in Section 6.

## 2 POLICY SEARCH USING STOCHASTIC GRADIENT ESTIMATION

We limit our attention to policy search methods, though there exist many other approaches to solving reinforcement learning problems [4]. Policy search methods typically perform hill climbing through a space of policies $\pi \in \Pi$. This section introduces a toy example and then describes a well-known method for estimating the policy gradient, i.e., the gradient of the expected total reward with respect to the policy parameters.

### 2.1 GENERAL SETTING, TOY EXAMPLE

In general, a control policy $\pi$ produces a sequence of control inputs $u_t$, driving the environment through a sequence of states $x_t$. The *history* of the system, $H$, is a random variable whose values $h$ are possible sequences of state–action pairs. A *response function* [5] $F(h)$ evaluates each actual history (typically by the sum of rewards at each time step). We seek an optimal policy $\pi^*$ that maximizes $E[F(H)]$ over the policy space $\Pi$, where the expectation is taken with respect to the distribution $P_\pi(h)$ over histories induced by $\pi$.

In our toy example (Figure 1), a cannonball is fired at a distant target. A *policy* $\pi = (\theta_d, v_d)^T \in \Pi$ consists of a *desired* cannon angle, $0 \le \theta_d \le \pi/2$, and *desired* initial velocity, $v_d > 0$. The control input $u_0$ is given directly by $\pi$ and the initial state $x_0$ consists of the *actual* velocity and angle, $(\theta_0, v_0)$. Thus, for this "one-shot" problem, the history is defined by just $h = (x_0, u_0)$, since the complete physical trajectory is determined by these values. The response function for this problem is defined to be $F(h) = -d(h)^2$, where $d(h)$ is the distance from the target to the point where the cannonball lands. Maximizing $E[F(H)]$ is equivalent to minimizing the expected squared distance error.

In the noise-free setting, the desired and actual values are identical. Furthermore, there is a continuum of values for $(\theta_d, v_d)$ that cause the ball to hit the target exactly, as shown in Figure 2.

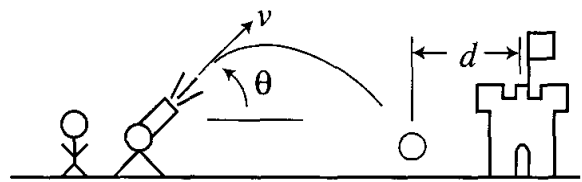

Figure 1: The cannon problem.

When there is noise, the actual velocity and angle imparted to the ball may differ from their intended values. Let the actual velocity and angle be $x_0 = u_0 + n_0$ where $n_0$ is a two-dimensional, zero-mean, Gaussian noise vector with covariance matrix $\Sigma$. Now, there is a *unique* optimal solution, as shown by the X in figure 2. This solution is at roughly $\theta_d = 45$ degrees, because this is the region where the contours are furthest apart (and hence the targeting is least sensitive to noise). The solution, is *not* on the noise-free optimal curve, however. The cannon should in fact be fired with a slightly higher velocity than that required in a noise-free environment because errors in the angle, whether positive or negative, will cause the ball to land short of the target.

### 2.2 STOCHASTIC GRADIENT ESTIMATION

We now describe a standard approach, due to Williams [10], for estimating the policy gradient from observed trials $(h^i, F(h^i))$ of the behavior of the policy and its response. From the trials, an estimate of the gradient $\nabla_\pi E[F(H)]$ is computed and the parameters of $\pi$ can then be adjusted in an attempt increase the expected response. We assume that a physical system or simulator draws samples from a known distribution (dependent on the policy $\pi$), and that we can measure the resulting control noise.[1]

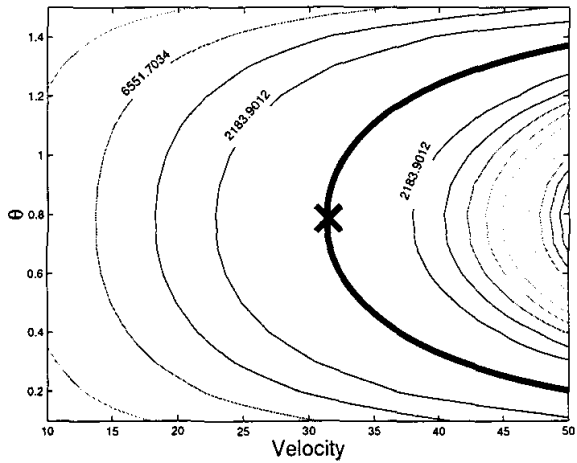

Figure 2: Squared error for the cannon problem. A contour plot showing the squared distance $d^2$ from the target as a function of $\pi = (\theta, v)$, assuming no noise. The solid black curve is the level set $d^2 = 0$, and represents the optimal noise-free solutions. The 'X' marks the optimal solution in a noisy environment and it lies slightly to the right of the noise-free solution curve.

We begin by writing out the expression for the gradient of the expected response, then move the gradient operator inside the expectation integral and rearrange to obtain an expression that has the form of an expectation with respect to $P_\pi(h)$:

$$\mathbf{v} = \nabla_\pi \mathrm{E}\left[F(H)\right] = \nabla_\pi \int F(h) P_\pi(h) dh$$
$$= \int \left[\frac{\nabla_\pi P_\pi(h)}{P_\pi(h)} F(h)\right] P_\pi(h) dh. \quad (1)$$

Given $N$ samples $(h^1, h^2, \ldots, h^N)$ drawn from $P_\pi(H)$, we can approximate (1) by

$$\widehat{\mathbf{v}}(h^1, \ldots, h^N) = \frac{1}{N} \sum_{i=1}^{N} \frac{\nabla_\pi P_\pi(h_i)}{P_\pi(h_i)} F(h^i). \quad (2)$$

To understand this equation, it will be helpful to define the *eligibility* of each sample point as follows:

$$\mathcal{E}(h) := \frac{\nabla_\pi P_\pi(h)}{P_\pi(h)} = \nabla_\pi \log P_\pi(h). \quad (3)$$

The eligibility measures how much the log likelihood of drawing a particular sample will change due to a change in $\pi$. The eligibility $\mathcal{E}(h)$ is a vector in policy space $\Pi$ that points in the direction of making $h$ more likely. For the cannon problem, recall that $h = (x_0, u_0)$. The eligibility of a particular cannon shot is $\mathcal{E}(h) = \Sigma^{-1}(x_0 - \pi)$. In other words, to make a history $h$ more likely, we should adjust $\pi$ to move in the direction $\Sigma^{-1} n_0$. Note that from (1) and (3), we have, for the true gradient,

$$\mathbf{v} = \mathrm{E}\left[\mathcal{E}(H) F(H)\right] \quad (4)$$

and, for the estimated gradient,

$$\widehat{\mathbf{v}}(h^1, \ldots, h^N) = \frac{1}{N} \sum_{i=1}^{N} \mathcal{E}(h^i) F(h^i). \quad (5)$$

---

[1]PEGASUS [6], to which we will compare our algorithm in Section 5, assumes complete control over the randomness introduced by the simulator.

Clearly, following such a gradient estimate will tend to adjust $\pi$ making the high-scoring histories more likely and the low-scoring policies less likely, as desired.

## 3 REDUCING GRADIENT ESTIMATE VARIANCE

When the policy is far from the optimal noise-free contour, the gradient estimate given by (5) tends to be quite similar to the noise-free gradient, and can be estimated from relatively few samples. On the other hand, in parts of the policy space near the noise-free optimal contour, the gradient signal is only reducing the effect of noise and is much fainter, requiring many more samples to estimate. This is especially true for high-dimensional problems. Because the cost of generating samples (whether simulated or physical) dominates the overall cost of motor control learning, this is a serious problem. We now discuss one existing and two new methods for reducing the variance of the gradient estimator and hence reducing the number of samples required.

Formally, these methods seek to reduce the trace of the covariance matrix, namely

$$\sigma^2 = \mathrm{Tr}(\mathrm{E}[(\widehat{\mathbf{v}} - \mathbf{v})(\widehat{\mathbf{v}} - \mathbf{v})^T])$$
$$= \mathrm{E}[(\widehat{\mathbf{v}} - \mathbf{v})^T (\widehat{\mathbf{v}} - \mathbf{v})].$$

By noting that the expected eligibility, $E[\mathcal{E}(H)]$, equals zero, we can construct a family of unbiased estimators by subtracting a constant *reinforcement baseline*, $a \in \mathbb{R}$

$$\widehat{\mathbf{v}}_a(h^1, \ldots, h^N) = \frac{1}{N} \sum_{i=1}^{N} \mathcal{E}(h^i) \big(F(h^i) - a\big). \quad (6)$$

The constant $a$ can, if judiciously chosen, reduce the variance of the estimator. In particular one can show

[9] that the minimal variance estimator of this form is obtained by setting

$$a = \frac{\mathrm{E}[\mathcal{E}(H)^T \mathcal{E}(H) F(H)]}{\mathrm{E}[\mathcal{E}(H)^T \mathcal{E}(H)]}. \quad (7)$$

While the above expression yields the minimal variance baseline estimator, its value is unknown and thus must be estimated from the samples. In practice, this will introduce estimator bias.

### 3.1 METHOD 1: FITTING THE RESPONSE SURFACE

One can treat the offset, $a$, in the baseline estimator as a simple model of $F$. Intuitively, the baseline term $a$ acts as a $0^{th}$ order prediction of $F$, i.e. $\widehat{F}(h) = a$. In the trivial scenario where the response surface is actually constant, the estimator $\widehat{F}$ can be used to compute the exact gradient regardless of the number of samples drawn (namely, the gradient would be zero everywhere). In more interesting cases in which the response surface is not constant, the variance will depend on how well the model $\widehat{F}$ predicts future outcomes.

In this paper we extend the idea of the reinforcement baseline by fitting a local linear in parameters (LIP) model of the response surface around the samples $(h_1, \ldots, h_N)$, namely

$$\widehat{F}(h) := \Phi(h)^T b. \quad (8)$$

where the *feature vector* $\Phi(h) \in \mathbb{R}^m$ is an arbitrary (possibly nonlinear) function of the designer's choice, and $b \in \mathbb{R}^m$ is the set of parameters to be fit. Define the unbiased, model-based estimator

$$\widehat{v}_b(h^1, \ldots, h^N) = G b + \frac{1}{N} \sum_{i=1}^{N} \mathcal{E}(h^i)(F(h^i) - \widehat{F}(h^i)) \quad (9)$$

where

$$G = \mathrm{E}[\mathcal{E}(H) \Phi(H)^T].$$

Intuitively, the function $\widehat{F}$ replaces the baseline $a$, in the previous estimator, and we also added the term $G b$ outside the summation. The term $G b$ is the stochastic gradient of the model itself. To see that the estimator is unbiased, note that

$$\mathrm{E}[\mathcal{E}(H) \widehat{F}(H)] = \mathrm{E}[\mathcal{E}(H) \Phi(H)^T] b = G b, \quad (10)$$

and therefore

$$\mathrm{E}[\widehat{v}_b] = G b + \mathrm{E}[\mathcal{E}(H) F(H)] - \mathrm{E}[\mathcal{E}(H) \widehat{F}(H)]$$
$$= G b + \mathbf{v} - G b = \mathbf{v}.$$

For many problems, such as the motor control problems considered in Section 4, $G$ can be computed analytically when using an appropriate choice of $\Phi$. Note that for the constant baseline model, $\Phi(h) \equiv 1$, and $b = a$, therefore the gradient predicted by the model equals 0.

The optimal linear response surface model $\widehat{F}$ is found by minimizing the variance of our estimator. The variance can be written as

$$N\sigma^2 = N\mathrm{E}[(\widehat{\mathbf{v}}_b - \mathbf{v})^T(\widehat{\mathbf{v}}_b - \mathbf{v})]$$
$$= \mathrm{E}[\|G b + \mathcal{E}(H)(F(H) - \widehat{F}(H))\|^2] - \|\mathbf{v}\|^2$$
$$= (b^T A b - 2 B^T b + C)$$

where

$$A = \mathrm{E}[\Phi(H)\Phi(H)^T \|\mathcal{E}(H)\|^2] - G^T G,$$
$$B = \mathrm{E}[\Phi(H) \|\mathcal{E}(H)\|^2 F(H)] - G^T \mathrm{E}[\mathcal{E}(H) F(H)],$$
$$C = \mathrm{E}[F(H)^2 \|\mathcal{E}(H)\|^2] - \|\mathbf{v}\|^2.$$

The minimal variance estimator is obtained by minimizing the above equation with respect to $b$. This turns out to be equivalent to shrinking the sum of the individual vector norms contributing to the gradient. The model that minimizes the variance of $\widehat{\mathbf{v}}_b$ is obtained by setting the derivative of the above expression with respect to $b$ to zero, and solving for $b$ to yield

$$b = A^{-1} B.$$

### 3.2 METHOD 2: WEIGHTED ELIGIBILITIES

In multi-step problems, one can view the gradient estimator as the sum of individual gradient estimators for each time step. This is due to the fact, that conditioned on a fixed policy $\pi$, the probability of generating a given history $h$ is given by a Markov chain. The eligibility can thus be factored as

$$\mathcal{E}(H) = \sum_{t=t_0}^{t_{\mathrm{final}}} \mathcal{E}^t(h), \quad (11\mathrm{a})$$

where

$$\mathcal{E}^t(H) = \frac{\nabla_\pi P_\pi(x_{t+1}|x_t)}{P_\pi(x_{t+1}|x_t)}. \quad (11\mathrm{b})$$

In certain settings (e.g., multiplicative noise), the variance of gradient estimators may be quite large. This is particularly problematic if the variance is high and the expectation itself is relatively small. Even in a single time step problem, such as the cannon problem, the variance of the gradient estimate with respect to each policy parameter $\pi_i$ may vary widely.

Recall that for the cannon example, the eligibility of a given cannon shot, $h = (x_0, u_0)$ is given by $\mathcal{E}(h) =$

$\Sigma^{-1}n_0$ (where $x_0 = u_0 + n_0$). Suppose that $n_0$ is a Gaussian with diagonal covariance matrix

$$\Sigma = \begin{bmatrix} \sigma_1{}^2 & 0 \\ 0 & \sigma_2{}^2 \end{bmatrix}.$$

If the noise in the angle $\theta_d$ is much less than that of the initial velocity $v_d$ ($\sigma_1 \ll \sigma_2$), then the variance of the derivative with respect to changes in $\theta_d$ will be tend to be higher. Intuitively, the variation in the response $F(h)$ is mostly due to variations in velocity since that is where most of the noise enters the system. However, there is no way to infer this from a single response value. This means that small fluctuations in control that are mostly deterministic will dominate the gradient estimation. This is counterintuitive because in most problems, these small variations give very little information about the gradient.

We propose to use a weighted version of our previous estimator. Others have used discounting to limit the contributions of past actions [1], the idea being that past actions have a smaller effect on the current reward signals obtained due to the mixing of the underlying Markov chain. Here, the weights will be used to mitigate the problem of having highly disparate variances. Consider the following gradient estimator:

$$\widehat{\mathbf{v}}_\Lambda = \Lambda \widehat{\mathbf{v}} \qquad (12)$$

where $\Lambda$ is a (fixed) weighting matrix, and $\widehat{\mathbf{v}}$ is an unbiased estimator, for example the estimator $\widehat{\mathbf{v}}_b$ introduced in the previous section. The motivation for using this form of an estimator comes from the fact that, in principle, a positive definite weighting matrix on the *gradient* will not imperil local convergence of a hill climbing algorithm. The matrix $\Lambda$ can be used to discount components of the gradient estimate that suffer from high variance; in particular, minimizing the mean-squared error

$$\text{MSE} = \mathrm{E}\big[\|\widehat{\mathbf{v}}_\Lambda - \mathbf{v}\|^2\big]$$

by restricting $\Lambda$ to be diagonal, with diagonal entries $\lambda_i$ given by

$$\lambda_i = \frac{N\mathbf{v}_i^2}{\mathrm{E}\big[(\widehat{\mathbf{v}}(H)_i - \mathbf{v}_i)^2\big] + N\mathbf{v}_i^2},$$

where $\widehat{\mathbf{v}}(H)$ is the "single sample" version of the estimator and subscript $i$ indexes the $i^{\text{th}}$ component of a vector.

Since the true gradient $\mathbf{v}$ is unknown, we can approximate the above equation by using the empirical estimate for the variance term and setting the $\mathbf{v}_i$ terms to some upper bound value $k$. This gives us the following equation:

$$\lambda_i = \frac{Nk^2}{\mathrm{E}\big[(\widehat{\mathbf{v}}(H)_i - \mathbf{v}_i)^2\big] + Nk^2}. \qquad (13)$$

If our upper bounds are correct, we will have an estimator that lies between the naïve estimate (with no discounting) and the optimal. Notice that as the number of samples $N$ approaches $\infty$, the above scaling term goes to 1.

## 4 MOTOR CONTROL

In this section we will show how to incorporate the ideas given in the previous section into a motor learning problem. Let $\mathcal{X}$ denote the state space of our system and the system's state $x_t \in \mathcal{X}$ evolves in discrete time ($I = \{t_0, t_1, \ldots, t_{\text{final}}\}$), for $t_0 < t_{\text{final}} \leq \infty$. Let $u_t \in \mathcal{U}$ denote the system's control at a given time instant. The system evolves according to:

$$\begin{aligned} x_{t+1} &= f(x_t, u_t + n_t) \\ y_t &= g(x_t) + w_t. \end{aligned} \qquad (14)$$

where $y_t \in \mathcal{Y}$ is the system output, available via some sensor suite, for example. We model the system as being corrupted by two noise processes: $n_t$ represents an *input noise* process and $w_t$ a *measurement noise* process.

The spaces $\mathcal{X}, \mathcal{U}, \mathcal{Y}$ are assumed to be, for simplicity, real vector spaces. Let $h = \{x_t, u_t\}_{t \in I} \subset \mathcal{X} \times \mathcal{U}$ describe the state-action history of our system.

A policy, $\pi : I \times \mathcal{Y} \to \mathcal{U}$, maps sensor values to controls. The explicit dependency on time of $\pi$ enables a spectrum of policies from "open loop" to "closed loop". Of course, buried in $\pi$ may be a state observer. For the current work, we restrict our attention to smooth systems $f$ and policies $\pi$ in the sense that $\partial f/\partial x$ and $\partial f/\partial u$ are well defined on $\mathcal{X} \times \mathcal{U}$, and $\partial \pi/\partial t$ and $\partial \pi/\partial y$ are well defined on $I \times \mathcal{Y}$.

We model the dynamics of our motor control tasks as a discrete time nonlinear system. We assume that at each time step, a controller generates a desired control signal $u_t$ that is then perturbed by Gaussian noise. This noise is centered around the desired control value and has covariance matrix $\Sigma(u_t)$. The dependency on the control value allows us to incorporate sources of multiplicative noise. The variance of the disturbed motor control noise $n_t$ can be written as

$$\Sigma(u_t) = \sum_{j=1}^{M} C_j u_t u_t^T C_j^T + \Sigma_0,$$

where matrices $C_j$ scale the Gaussian noise.

## 4.1 OPEN LOOP CONTROL

An open loop controller consists of a trajectory of control values $u_t$ that are fixed in advance. These control signals are then executed without any feedback from the environment. To calculate the eligibility for a given history $h$, we will need to calculate how each control signal varies with respect to $\pi$ at each time step.

For example, one representation of an input trajectory is a spline where the policy parameters $\pi_i$ control the placement of knot positions at fixed time intervals. Since the value of a spline at time $t$ is a linear function of the knot positions, this derivative can easily be computed.

## 4.2 TRAJECTORY TRACKING WITH PD CONTROL

A proportional derivative (PD) controller is one that uses a simple form of feedback to correct for errors from some desired path. The state $x_t = (q_t^T, v_t^T)^T$ includes the positions and velocities of the system. The control value $u_t$ is proportional to the difference between the state $x_t$ and some desired state $x_t^* = (q_t^{*,T} v_t^{*T})^T$. The control signal at time $t$ is

$$u_t = K(x_t^* - x_t)$$

where the gain matrix $K$ is assumed fixed. To compute how the control signal changes with respect to a change in our desired state $x_t^*$, apply the chain rule to obtain

$$\frac{\partial u_t}{\partial \pi_{(i)}} = \frac{\partial u_t}{\partial x_t^*}\frac{\partial x_t^*}{\partial \pi_{(i)}} = K\frac{\partial x_t^*}{\partial \pi_{(i)}}. \quad (15)$$

## 4.3 THE ELIGIBILITY

The gradient estimators require the computation of the eligibility. The probability of drawing a particular sample $h$ is given by the following equation:

$$P_\pi(n_t|x_t) = \frac{1}{(2\pi)^{d/2}|\Sigma(u_t)|^{1/2}} \cdot e^{(-\frac{1}{2}n_t^T \Sigma(u_t)^{-1} n_t)}$$
$$P_\pi(h) = \prod_{t=1}^{T_f} P_\pi(n_t|x_t). \quad (16)$$

Substituting (16) into (11) we have

$$\mathcal{E}^t(h)_{(i)} = -\frac{1}{2}\text{Tr}\left(\Sigma(u_t)^{-1}\frac{\partial \Sigma(u_t)}{\partial \pi_{(i)}}\right)$$
$$+ n_t^T \Sigma(u_t)^{-1}\frac{\partial u_t}{\partial \pi_{(i)}}$$
$$+ \frac{1}{2}n_t^T \Sigma(u_t)^{-1}\frac{\partial \Sigma(u_t)}{\partial \pi_{(i)}}\Sigma(u_t)^{-1} n_t,$$

where

$$\frac{\partial \Sigma(u_t)}{\partial \pi_{(i)}} = \sum_{j=1}^{M} C_j\left(u_t\frac{\partial u_t^T}{\partial \pi_{(i)}} + \frac{\partial u_t}{\partial \pi_{(i)}}u_t^T\right)C_j^T,$$

and the terms $\partial u_t/\partial \pi_{(i)}$ are given by (15).

## 5 EXPERIMENTS

We demonstrate our algorithm by finding an optimal policy $\pi^*$ for a dart-throwing task. The objective is to throw a dart with minimal mean squared error (measured from where the dart hits the wall to the center of the dart board). The arm is modeled as a three-link rigid body with dimensions based on biological measurements [2]. The links correspond to the upper arm, forearm, and hand. These are connected to each other using a single degree of freedom rotational joint and the upper arm is connected to the shoulder at a fixed location. We generated code to simulate the dynamics of this system using SD/Fast, a software package for physically based simulations.

The arm is controlled by applying a torque at each joint. These torques are generated by a PD-controller that attempts to move the arm through a desired trajectory, specified by a cubic spline for each joint. The starting posture of the arm is fixed in advanced and the path is determined by interpolating between three other knot positions. These three knots per joint give us a compact policy representation of 9 parameters. The controller is simulated for approximately 0.2 seconds and then the dart is released (there is Gaussian noise added to the release time with $\sigma = 0.01$). Additional noise enters the system by perturbing the torques $u_t$ given by the PD-controller by additive and multiplicative noise.

We implemented the ideas presented in section 3.1 by choosing an appropriate feature map $\Phi$ and calculating the gradient of the corresponding response surface model. At each hill climbing step, multiple samples are drawn and used to estimate the optimal feature weights $b$. We originally tried a mapping that included terms for the sum of the noise signals $n_t$: $\Phi(h) = [1 \quad \sum_{t=t_0}^{t_{\text{final}}} n_t^T]^T$. If the arm produces more torque in one joint than expected over a sample trajectory, then this difference may correlate well with the response. The gradient according to the model is $G = [0 \quad \sum_{t=t_0}^{t_{\text{final}}} \partial u_t/\partial \pi]$. This appeared to give improvements for situations where the release time was fixed. However it did not improve performance after we added noise in the release time. Instead we found that using the release time $t_r$ and $t_r^2$ as features,

$$\Phi(h) = [1 \quad t_r \quad t_r^2]^T, \quad (17)$$

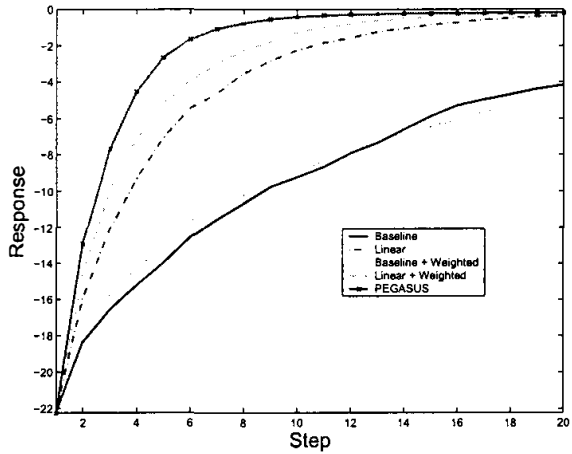

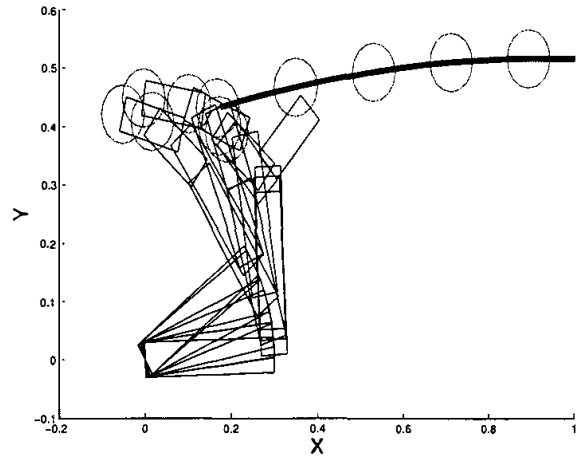

Figure 3: Learning curve for the dart-throwing problem. This graph is averaged over 100 hill climbing episodes with 100 samples drawn at each step. The linear response surface model gives a significant improvement to the rate of convergence over the optimal baseline. Using weighted eligibilities gives a further improvement for the linear model. PEGASUS, which makes stronger assumptions about the simulation environment, outperforms both methods.

Figure 4: Superimposed images of our agent trying to hit the bullseye. The solid line shows the path of the dart after it is released.

## 6 CONCLUSIONS

Figure 4 shows one trial of a locally optimal policy for the dart thrower. The motions generated by this policy are, to a human observer, extremely natural. (See www.cs.berkeley.edu/~gregl/uai03-videos/ for examples.) This lends support to Wolpert's claim that noise is a major factor in determining biological motor control strategies. Notice in particular that the path of the hand prior to the dart release follows the trajectory of the dart as if it were already in free flight. This strategy minimizes the error introduced by noise in the release time. In general, we expect that, perhaps counterintuitively, injection of noise into physically-based animation is likely to result in much more physically realistic motion behavior.

yields a considerable reduction in the variance of the gradient estimator, and thus improved hill-climbing performance. Since the release time is independent of the policy parameters and the expected eligibility is zero for any release time, the gradient according to the model is zero ($G = 0$).

Figure 3 shows the learning curve for the dart-throwing problem. This graph is averaged over 100 hill climbing episodes with 100 samples drawn at each step. The best response seen so far is plotted at each hill climbing step (some episodes diverged in our experiments). The linear response surface model gives a significant improvement to the rate of convergence over the optimal baseline. Using weighted eligibilities to reduce the effects of high variance components appears to give further improvements in the gradient estimates for the linear response surface model. The reweighted eligibilities do not appear to improve the baseline results. In both cases, an upper bound was placed on the squared gradient for each vector component ($k^2 = 10$).

The PEGASUS curve was generated by using a finite difference method to estimate the gradient. This method outperforms the other techniques, but makes stronger assumptions about the simulation environment. The other gradient estimates could in principle be implemented on a real system, provided that there is a way to measure the noise.

Both of the variance reduction techniques we have introduced require estimating parameters that are then incorporated into the basic gradient estimator equation (4). While the optimal parameter equations are exact in expectation, estimating these values may prove to be difficult. In fact, if one is not careful, this procedure could cause the gradient estimates to suffer from higher variance. In general, one should draw more samples to fit models that are more complex. Sample reuse [7] could limit this problem in some situations, improving performance.

Whereas we assumed full observability, others have presented gradient estimation techniques that apply in partially observable domains. These techniques typically assume uncertainty in the state variable, but complete access to the disturbed control signal $u_t + n_t$. We would like to explore the case where the controller receives a noisy measurement of the disturbance $n_t$ as we believe a solution to this problem will lead us a step further to being able to design learning algorithms appropriate for real physical systems.

PEGASUS reduces the variance of the gradient estimate by reusing the same noise signals when evaluating different policies. This assumes access to a simulator that can produce samples with fixed perturbations. We would like to explore how this idea can be used to improve our algorithm in the simulation environment. One possible extension would involve sampling points under a proposal distribution, different from $P_\pi(H)$.

**Acknowledgments**

This research was supported by NSF/KDI Grant FD98-73086 and by ARO MURI Grant DAAD19-02-1-0383. We would like to thank Peter Bartlett for discussions on gradient estimation techniques applied to reinforcement learning.